\documentclass[final,3p,times,twocolumn]{elsarticle}

\usepackage{amssymb}
\usepackage{algorithmic}
\usepackage{graphicx}
\usepackage{textcomp}
\usepackage{tabulary}
\usepackage{multirow}
\usepackage[]{xcolor}
\usepackage[para]{threeparttable}
\usepackage{array,booktabs,longtable,tabularx}

\usepackage{caption}
\usepackage{subcaption}

\journal{Engineering Applications of Artificial Intelligence}

\begin{document}

\begin{frontmatter}



\title{A Machine Learning Approach to Classifying Construction Cost Documents into the International Construction Measurement Standard}


\author[1]{J. Ignacio Deza\corref{a}}

\author[2]{Hisham Ihshaish\corref{a}}

\author[1]{Lamine Mahdjoubi}

\address[1]{Centre for Architecture and Built Environment Research (CABER).}
\address[2]{Computer Science Research Centre (CSRC) and Mathematics and Statistics Research Group (MSRG).}
\address{Faculty of Environment and Technology, University of the West of England, Bristol, BS16 1QY, United Kingdom}
\cortext[a]{Corresponding Authors: J. I. Deza and H. Ihshaish (e-mail: \{ignacio.deza, hisham.ihshaish\}@uwe.ac.uk). }

\begin{abstract}

We introduce the first automated models for classifying natural language descriptions provided in cost documents called ``Bills of Quantities'' (BoQs) popular in the infrastructure construction industry, into the International Construction Measurement Standard (ICMS). The presented analysis and models are aimed at vitalising the adoption of ICMS and thus providing benchmarkers with an effective automated tool to allow for project comparison in a more granular way. The presented study addresses these challenges and sets forth models to facilitate widespread analysis of cost and performance in infrastructure construction projects effectively. The models we deployed and systematically evaluated for multi-class text classification are learnt from a dataset of more than 50 thousand descriptions of items retrieved from 24 large infrastructure construction projects across the United Kingdom. 

We describe our approach to language representation and subsequent modelling to examine the strength of contextual semantics and temporal dependency of language used in construction project documentation. To do that we evaluate two experimental pipelines to inferring ICMS codes from text, on the basis of two different language representation models and a range of state-of-the-art sequence-based classification methods, including recurrent and convolutional neural network architectures.  

The findings indicate a highly effective and accurate ICMS automation model is within reach, with reported accuracy results above $\%90$ F1 score on average, on 32 ICMS categories.  Furthermore, due to the specific nature of language use in the BoQs text; short, largely descriptive and technical, we find that simpler models compare favourably to achieving higher accuracy results. Our analysis suggest that information is more likely embedded in local key features in the descriptive text, which explains why a simpler generic temporal convolutional network (TCN) exhibits comparable memory to recurrent architectures with the same capacity, and subsequently outperforms these at this task.
\end{abstract}


\end{frontmatter}


\section{Introduction}

One of the biggest challenges the construction industry is facing worldwide is \textit{standardisation}. Compared to other big industries (manufacturing, software, financial and medical services being just a few paradigmatic examples), construction projects are still lagging behind. Their handling remains largely as a craft,  reminiscent of the pre- Henry Ford assembly-line manufacturing era, where each car was slightly different, done in slightly different ways and with nearly ten times more effort than a single standardised vehicle\cite{thompson2017digitalisation,davies2018transform}. The impact is real. Globally, 98\% of infrastructure projects are over budget or delayed, with an average of 80\% over budget and at least 20 months late. Construction’s productivity is also lagging global productivity by over 30\%.  \newpage If the productivity of the construction industry matched average global productivity, it would pay for 50\% of the total demand of infrastructure\cite{changali2015construction}. At the end of the road to standardisation lies a pot of gold worth billions.

There can be several causes to this phenomenon: firstly, whereas manufacturing is done in a controlled environment where pieces can be consistently made to specifications, construction (in the brick-and-mortar sense) is usually done on site and it is subject to the weather, logistics, terrain problems and other issues\cite{pan2008leading}; secondly, whereas a company is usually responsible for the production of a good or service, in construction many companies and subcontractors across many trades have their responsibilities intertwined which results in liability issues when under-performance occurs\cite{fewings2019construction}; thirdly, construction tends to employ unskilled workers -- not necessarily being trained to perform processes that are repeatable, replicable, and linear in nature - whose performance tends to be harder to measure\cite{fewings2019construction,davies2018transform}; and lastly, data recollection in other industries usually follows certain standards and specifications which are cross discipline and cross country, enabling e.g. international commerce, and ensuring predictability and wide compatibility with other products, projects and applications; in the construction industry, nearly every company records their data in slightly different and usually proprietary ways, catered to their own specific needs.

The first three issues are being addressed by the use of new technologies like BIM\cite{yin2019building,thompson2017digitalisation}, digital twins\cite{pan2008leading,el2020digital}, and especially by the rising use of \textit{Offsite Construction}\cite{pan2008leading,yin2019building}, which brings new technologies to the construction site by manufacturing standardised parts in an assembly line and shipping them only to be assembled on site. This technique is gaining popularity due to its more predictable nature, their use of standardised processes and their repeatability. The fourth issue is---however---more challenging, as it requires for the whole industry to come together around a standard. 

Materials and work in the construction industry are usually billed using documents called Bills of Quantities (BoQs)\cite{fewings2019construction} which contain a free text (natural language) description of the material or work conducted together with a price breakdown. This is one of the elementary types of billing document and many types of contracts in construction (i.e. with sub-contractors, etc\ldots) can be traced back to BoQs. Some of these documents may also include a code which defines the type of work, or the material. Unfortunately, these codes lack standardisation and tend to be internal to the company, to the industry---usually reflecting industry needs like tracks and electrification for train infrastructure---and in some situations, to the country, where a standard set of measurement rules has been imposed, like the NRM standard\cite{wu2014can} used in the UK for the construction of buildings.

\begin{figure*}[h!]
  \includegraphics[width=\linewidth]{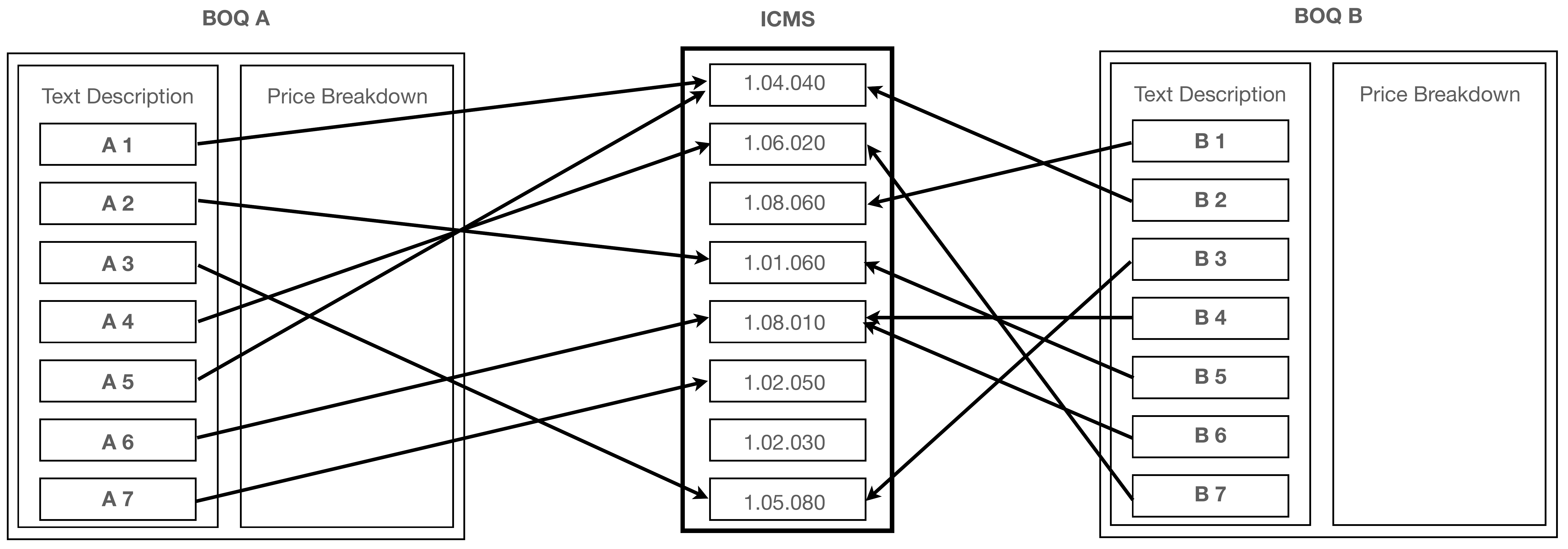}
  \caption{ICMS is designed to allow comparison between BoQ-like documents which--in general--don't have means of comparison apart from the text description. Although many very detailed rules of measurement and naming conventions exist, they tend to be so granular that the comparison between items becomes harder because similar items fall in different categories. For this reason, ICMS is designed as a high-level, elementary cost and carbon measurement standard, with the potential to make cost documents more transparent and international.}
  \label{fig:BOQ}
\end{figure*}

Due to the aforementioned reasons, it has been very difficult to establish and adopt a universal standard, as most contenders are country-specific or industry-specific. Usually such standards are designed to be highly granular, with hundreds or even thousands of categories, which makes it all the more challenging to compare like-for-like when benchmarking BoQs from different origins. For this reason there has been an effort from a group of 49 professional and not-for-profit organisations from around the world, working towards developing and implementing an international standard for benchmarking, measuring and reporting construction project costs. This group created the International Cost Management Standard (ICMS).

The ICMS\cite{muse2021icms, mitchell2016international} aims to provide global consistency in classifying, defining, measuring, analysing and presenting entire construction costs at a project, regional, state, national or international level.
ICMS is a cost classification system, and contrary to many word-break structures, the project has global coverage, with the scope of standardising entire construction cost documents.
Its purpose is not to replace other--more specific--methods of measurement, but to complement them. It is designed in an elementary way, and--by design--it is not highly granular, thus allowing different projects from different sectors (and different countries) to be comparable at that level.  

ICMS is focused on many aspects of construction projects: capital costs (as standardisation brings significant benefits to construction cost management), life cycle costs (reflecting the pivotal role of financial management in construction), and also ‘carbon emissions’ (carbon dioxide ($CO_2$) equivalent) accountable in a way akin to monetary cost\cite{deo2019race}.

The standard has been successfully implemented by many bodies, including the Africa Association of Quantity Surveyors (AAQS), the China Cost Engineering Association (CCEA), the European Federation of Engineering Consultancy Associations (EFCA), the International Cost Engineering Council (ICEC), and the Royal Institution of Chartered Surveyors (RICS), among others. It is currently the sole contender to become a global standard to allow a meaningful comparative analysis inside and between countries, by international organisations such as the World Bank Group, the International Monetary Fund, various regional development banks, non-governmental organisations and the United Nations\cite{mitchell2016international,muse2021icms}. 

For these reasons the ICMS project presents itself as a very high profile venture, with the potential to disrupt the present construction methodologies towards a more transparent, international and sustainable future.

The work presented in this article aims to foster \textit{adoption} of this new standard. Adoption of a common standard is usually difficult unless there is immediate gain by the participants. As with many processes in the construction industry, classification of BoQs is usually done manually. This makes adoption of an extra standard---focused on benchmarking and  optimisation, rather than on day-to-day operations---an extra burden that more often than not will tend to be avoided. Fortunately, the Department for Transport (DfT)\footnote{https://www.gov.uk/government/organisations/department-for-transport} of the UK has directed many Government-owned-Companies who administer parts of its infrastructure to comply with the ICMS. This study, which comes under the TIES Living Labs\cite{ties2022} project sponsored by the UK Government, is in line with these efforts.

\section{Data and Methods}

The natural language descriptions found in the BoQs are predominantly short. Similar to texts analysed in studies of sentiment analysis\cite{li2014news}, dialogue systems\cite{lee2016sequential, ryandialogue, Reece} and user query intent understanding\cite{hu2009understanding}, among others, inferring such text is known to be especially challenging. This is because of the often limited contextual information they are accompanied by, compared to that of long texts found in books and documents.

 The contextual information in the analysed texts is naturally present, albeit simpler, compared to the complexity often embedded in other types of natural language. This is largely due to the fact that BoQs descriptions are considerably condensed, short and strictly descriptive, as they are essentially intended to be informative; with neither emotional nor opinion components to them. This can add to the challenge of extracting the semantics for the classification effort compared to other types of short texts from media or tweets.
 
Moreover, the classification of the description of tasks in natural language into a set of categories requires a certain degree of interpretation. This problem aggravates when there are multiple people performing the classifications and if the description lacks proper context. For example, all the load bearing works underground or underwater must be classified as ``substructure'', but when they are over ground they become ``structure''. There are however many structures that can be partially buried due to terrain issues, slopes, etc\ldots so the classification of such items may well be down to subjective judgment. This way a great variety of slightly different classifications can occur naturally in manual classification, which accentuates the need for an impartial classifier which can resolve these issues in an objective way. As the purpose of this standard is to compare like-for-like, even a systematically erroneous classification is desirable over classifications around a theme which won't match most items as intended. 
 
In this section we describe the dataset and the modeling methods we applied to automate the classification of BoQs into the ICMS standard.

\subsection{Data acquisition and pre-processing}

As a part of a the TIES Living Lab, a total of ~124 thousand Materials and Costs items, defined in natural language and originally labelled manually in ICMS, where retrieved from a total of 24 projects from a major UK-based infrastructure construction company. 

The data is presented as a cost documents, which include the ICMS code, the free text description to be analysed and a price breakdown per item. The prices were not considered in the study. 

Each project contains several thousand lines of cost descriptions,  written in natural language presumably by subcontractors executing the tasks or delivering the materials along the supply chain. These pieces of text are relatively short (the median length of the descriptions in the dataset is 14 words with a maximum of 160 and a minimum of only one word). The encoding (each sample mapped to an ICMS code) has been performed manually by Quantity Surveyor Experts with the support of the (British) Royal Institution of Chartered Surveyors (RICS\footnote{https://www.rics.org/uk/}), which is part of the ICMS Coalition which aims at promoting a widespread adoption of ICMS as a global standard. 

\begin{figure*}[!ht]
  \includegraphics[width=0.95\linewidth]{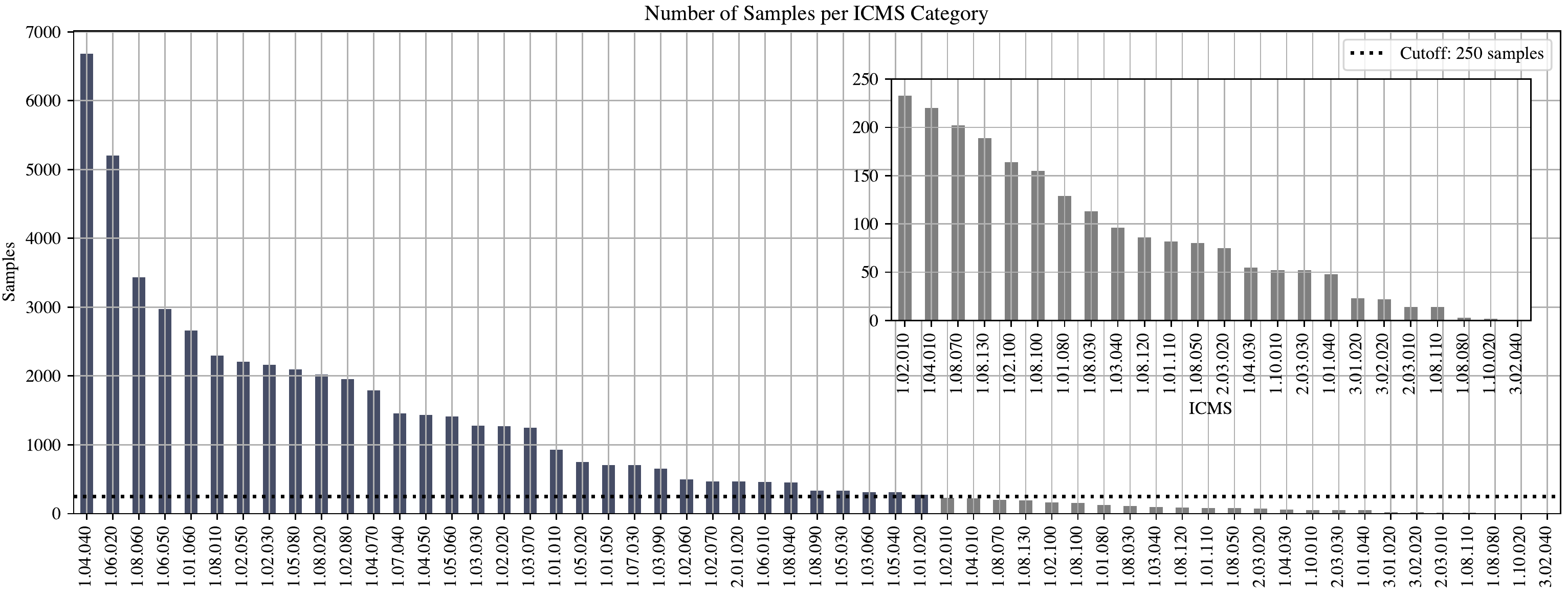}
  \caption{A histogram of the dataset; sample numbers per ICMS codes showing data imbalance. A cutoff of 250 is applied to exclude the overly under-represented ICMS codes from the analysis. (inset) The categories below the threshold and thus not included in the study. }
  \label{fig:histoICMS}
\end{figure*}

The number of unique ICMS categories with at least one entry in the original dataset is 72 (from the overall total of 109 categories present in the cost side of the standard). However, many of these categories contained only a handful of items and were as a result discarded in this study.  Only 32 categories contained \textit{sufficient} samples for their use in the presented study, reducing the total size of the dataset from 123210 to 51906 items --having additionally removed duplicated samples. We established a cut-off of 250 samples per ICMS category such that all ICMS categories with less number of samples were removed from the dataset. The distribution of items/samples per each ICMS encoding is provided in Fig. \ref{fig:histoICMS}.

The following synthetic examples bear a very close resemblance of the text analysed in this study, both in wording and in length -- the original data is protected by a non-disclosure agreement, and therefore cannot be shared publicly at this stage:

\begin{itemize}
\item ``Galvanised high adherence reinforcing strips acting as soil reinforcement.''
\item ``Take down and remove to tip off Site unlit traffic sign including 4 posts.''
\item ``Installation of wildlife tunnel XX m in length as per diagram XX.''
\item ``Geophysical Survey in accordance with drawing XX.''
\item ``Termination of optic fibre cable to XX equipment cabinet Type YY.''
\item ``Power reduction joint of XX mm2 to XX mm2.''
\end{itemize}

Since the retrieved descriptions have been recorded by a large number of different people, varying levels of complexities inherent in natural language were present as a result. We found distinguishable differences in detail level provided across data samples---e.g. references to internal codes, drawings, diagrams, scales, etc\ldots The recording  is additionally found in many inconsistent ways, i.e. `cable 10 m', `cable 10 meter', `cable 10m', `ten metre cable' which essentially are intended to record the same information. Additionally many words have been found misspelled in at least one way.

Such inconsistencies were considered as data was cleansed, and special characters---including punctuation---and numbers were removed. 

\subsection{Classification Methods}

For language representation and subsequently modelling, we considered two different approaches; an explicit representation for text with a vector space model\cite{SALTON1988513} based on term(s) occurrence, and an implicit representation, using a word embedding approach\cite{word-embedding} to text so that contextual semantics, beyond term occurrence, are represented to learn the corresponding \textit{target} labels.

For term occurrence models we used the popular \textit{n}-gram “bag-of-words” (BoW) whereby each unique term (or $n$  terms) is considered as an independent dimension of the terms space, and is “one-hot” encoded as a sparse vector.  Different weightings for term occurrence were additionally evaluated; a binary one-hot encoding, term frequency and the popular Term Frequency-Inverse Document Frequency (TF-IDF)\cite{tf-idf}. Although popular, allowing models to learn corresponding targets based on local key features, this approach is nonetheless limited as it considers terms in text to be independent, and as a result the semantic term-term dependence is entirely undermined.

On the other hand, \textit{word vectors}\cite{word-embedding}, also known as word embeddings, provide a much more semantic-aware representation for language, where each word (in the vocabulary, $V$) is embedded into a real-valued vector in a dense space of \textit{concepts}, of dimension $d << |V|$. 

\vspace{2cm}

The $d$ dimensions encode concepts shared by all words, rather than a statistic relative to each unique word. This generally allows for richer word-word relationship information than language representation of BoW models . Word vectors  may either be initialized randomly and trained along with machine learning models on a specific text mining task, or can be pre-trained vectors. We evaluated both approaches, and used a pre-trained word2vec\cite{word2vec} for word embeddings. 

To learn ICMS classes from possible contextual information provided in the BoQ, we further evaluated a set of deep learning methods as shown in \ref{fig:Ng2}, most widely used as sequence processing models. In particular we evaluated two RNN (recurrent neural network) architectures; the Bidirectional LSTM, or BiLSTM, a bidirectional RNN consisting of a forward LSTM\cite{LSTM} unit and backward LSTM unit to enhance the ability of neural networks to capture context information, and a simpler model of BiGRU\cite{biGRU}, or bidirectional gated unit consisting of the output state connection layer of forward GRU\cite{GRU}, reverse GRU, and forward and reverse GRU. Both models make up for the basic RNN architecture — which additionally are known to be notoriously difficult to train\cite{RNNBengio, RNNDifficult} — in extracting global features from text sequence, and have been widely used with notable improvement over basic LSTM and GRU architectures in a range of applications to text and speech processing\cite{shorttextLSTM, BahdanauCB14, bicomparison, gruspoken, bilstmChinese}.

We additionally evaluated a convolutional neural network (CNN)\cite{LeCun}, which has been applied to model sequences for decades, and more recently at tasks for text classification---e.g., sentence classification\cite{ConSentence1, ConvSentence2}, document classification\cite{condoc1, condoc2, condoc3} and sentiment analysis\cite{consentiment1, consentiment2}. ConvNets utilises multiple convolution kernels of different sizes to extract key information in sentences, which can capture the local relevance of text. A variant of CNNs, namely the Temporal Convolutional Network (TCN) was recently proposed and has shown promising performance results over standard CNN and RNN architectures on different NLP benchmarks\cite{TCN}. We evaluated a TCN architecture, primarily as these---contrary to CNNs which can only work with fixed-size text inputs and usually focus on terms that are in immediate proximity due to their static convolutional filter size---applies techniques like multiple layers of dilated convolutions and padding of input sequences in order to handle different sequence lengths and capture dependencies between terms that are not necessarily adjacent, but instead are positioned on different places in a sequence. This could potentially emphasise the strength of signal which can be dispersed in a given BoQ sequence. E.g. (from previous examples): ``\textbf{Take down} and \textbf{remove} to tip off Site unlit \textbf{traffic sign} including 4 posts.'', regardless of terms' proximity.

\subsubsection{Experiments and Model Description}

Associating the samples provided in the BoQs to ICMS categories can be learned by machine learning methods for classification, casting the task as a result to a supervised learning problem for natural language processing. For the studied dataset, $S = \{s_1, s_2, s_3, ..., s_n\}$, where $s_1, s_2, ..., s_n$ are the short texts provided in the independent BoQs and $|{S}| = 51906$, the corresponding ICMS categories $y_1, y_2, y_3, ..., y_m$ are provided as ground-truth labels. Each BoQ item, $s_i$, is associated to a unique ICMS category, $y_i \in{Y}$, where $|Y|$ = 32.

\begin{figure}
     \centering
     \begin{subfigure}[b]{0.5\textwidth}
         \centering
         \includegraphics[width=\textwidth]{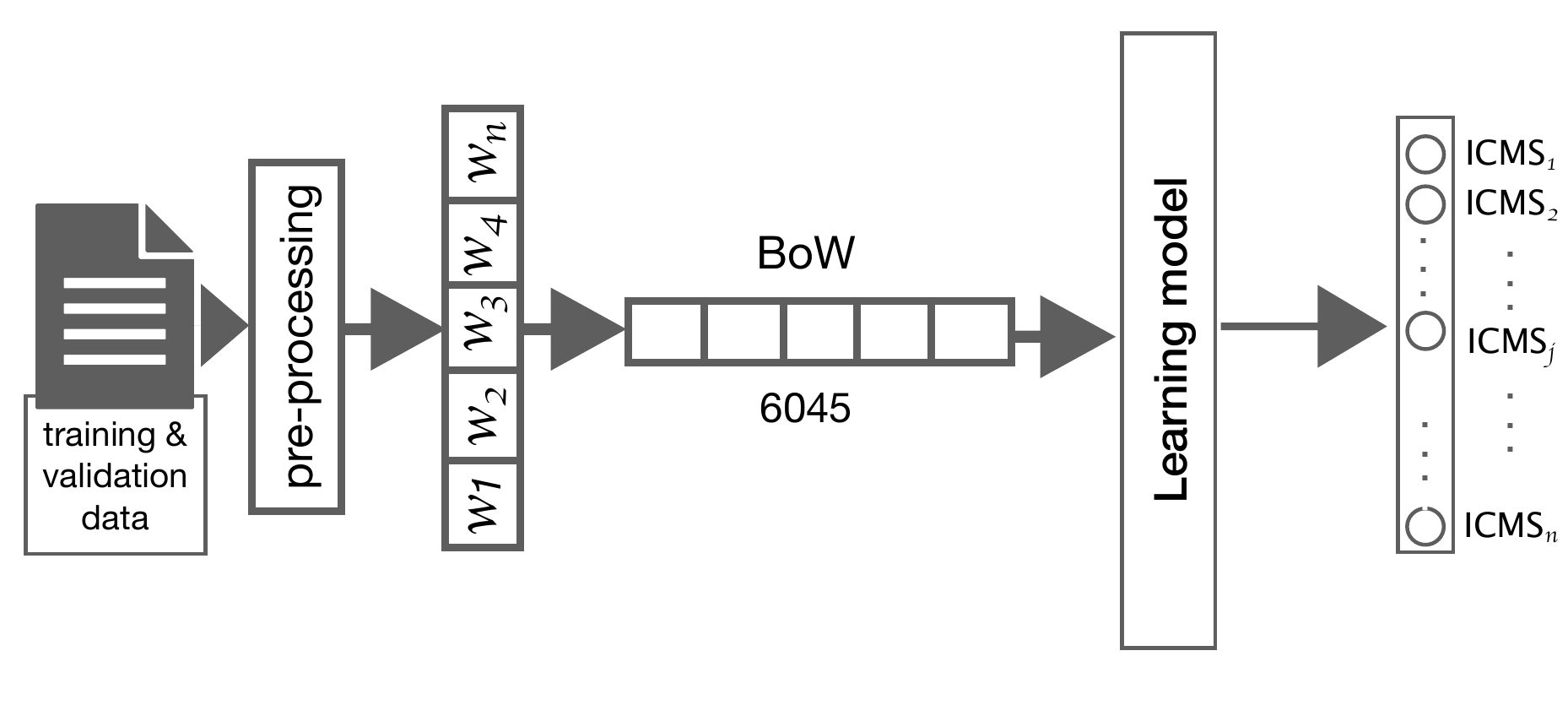}
         \caption{Pipeline 1 examining the performance of different classification models with bag-of-words for language representation.}
         \label{fig:Ng1}
     \end{subfigure}
     \hfill
     \begin{subfigure}[b]{0.5\textwidth}
         \centering
         \includegraphics[width=\textwidth]{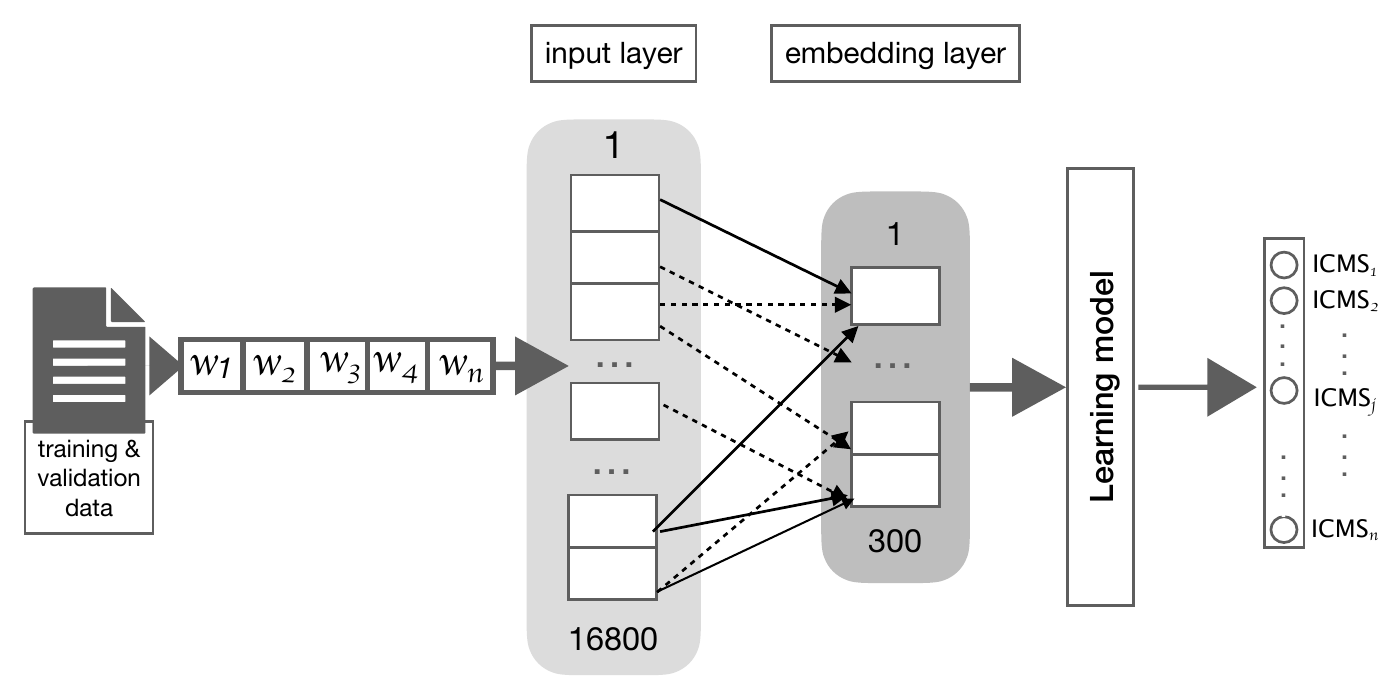}
         \caption{Pipeline 2 examining the performance of deep learning models with a word embedding layer.}
         \label{fig:Ng2}
     \end{subfigure}

        \caption{Schematic view of the two modeling pipelines; (a) with \textit{uni}-gram representation of the BoW model. The total number of unique terms is reduced to 6045 after applying stemming, lemmatization and stop-word removal. (b) an embedding layer is applied to learn a word representation of 300 dimensions ($d=300$ and $|V| = 16800$), or otherwise using a pre-trained word-embedding (see details on the the use of Word2Vec).}
        \label{schematic}
\end{figure}

To evaluate how the performance of classification models based on both language representation approaches compare, especially given the unique characteristics of language use in the BoQs; short (of uneven size), application-specific and predominantly descriptive, we evaluated classification methods in two different experimental settings as shown in Fig. \ref{schematic}.

In \textbf{Pipeline 1}, we trained and fine-tuned Support Vector Machine (SVM)\footnotemark \cite{svm}, Random Forest\footnotemark[\value{footnote}]\footnotetext{The model was built using Scikit-learn: scikit-learn.org}\cite{RF1, RF2} and multilayer perceptron (MLP)\footnote{The model was built using TensorFlow: tensorflow.org} algorithms.

The MLP --fully connected feed-forward neural network-- is trained with an input layer of size 300, an additional 50-sized hidden layer and a softmax output layer of 32.

In \textbf{Pipeline 2} we evaluated the three models of BiLSTM, BiGRU and TCN, each applying two different word embeddings as in\cite{word2vec}; learned vectors in an embedding layer with random initial weights, or pre-trained Word2Vec embeddings (trained on the Google News corpus  containing 100 billion words)\footnote{https://code.google.com/p/word2vec/}. We use a 300-dimension word embedding for both the learned and pre-trained. The learning process therefore consists of first, the semantic representation of each text is obtained through the model training (or using pre-trained skip-gram model), and the vector representation of words is obtained. Subsequently the vector representation of the word is input into each model for further analysis and extraction of semantics. The final word vector is then connected to a softmax layer of size 32 for text classification.

For all neural network models, including the MLP, we use ADAM\cite{adam} for learning, with a learning rate of 0.01. The batch size is set to 64. The training epochs are set to 40.
We employed the BiLSTM model as in\cite{biLSTM}, with two hidden layers of size 64. The BiGRU model is similarly trained with two hidden layers of the same size. A dropout rate of 0.5 is applied to both. All models in Pipeline 2 were implemented using Keras \footnote{F. Chollet. Keras. https://github.com/fchollet/ keras, 2015} and TensorFlow \footnote{Software available from tensorflow.org}.

The TCN uses 1D CNN layer, followed by two layers of dilated 1D convolution. We apply an exponential dilation $d = 2^i$ for layer $i$ in the network. Acasual convolutions are applied in the TCN so that target labels can be learnt as a function of terms at any time step in the sequence --contrary to the causal convolution that is used in Wavenet\cite{wavenet}-- with kernel size set to 3 where each layer uses 100 filters.

The dataset is split into a training and validation set (development set) of 80\% of the entire corpus to train and fine-tune the models, and a test set of 20\% (resulting in 10242 samples) to evaluate the different models' performance. The models were fine-tuned optimising the categorical cross-entropy $l = -\sum_{c=1}^{C=32}y_{s_i,c}\log(p_{s,c})$, where $p$ is the predicted probability observation $s$ is of class $c$, of total 32 ICMS classes in the dataset.

\subsection{Results and Analysis}

We used TF-IDF with a Multinomial Naïve Bayes\cite{multinomialNB} as a baseline model. Count Vectorizer and \textit{uni}-gram model with feature set size of 6045 was used. Similarly, the CNN of\cite{ConvSentence2} --a classic baseline for text classification -- based on the pre-trained word embedding is additionally used as a baseline model.  \newpage A synopsis of the results on both the accuracy and the macro F1 score for all the presented models on the test set is presented in Table \ref{table:results}. 

\begin{table} 
\footnotesize
\begin{tabular}{clp{1.4cm}cc}
\toprule
\multicolumn{1}{c}{} \textbf{Pipeline} & Model & Embedding & Accuracy & Macro F1 \\
\cmidrule(l){2-5}

\multirow{4}{*}{\rotatebox{45}{\textbf{Pipeline 1 }}}

&\textit{NB} & BoW & 0.861 & 0.857\\
&RF & BoW & 0.922 & 0.918 \\
&SVM & BoW & 0.863 & 0.860 \\
&MLP & BoW & \textbf{0.932} & \textbf{0.930} \\ 

\cmidrule(l){2-5}
 
\multirow{7}{*}{\rotatebox{45}{\textbf{Pipeline 2}}}
&\textit{CNN} & pre-trained & 0.898 & 0.837 \\
&BiGRU & pre-trained & 0.911 & 0.907 \\
&BiGRU & trained  & 0.926 & 0.913 \\
&BiLSTM & pre-trained & 0.903 & 0.898 \\
&BiLSTM & trained  & 0.919 & 0.907 \\
&TCN & pre-trained & 0.915 & 0.902   \\
&TCN & trained  & \textbf{0.929} & \textbf{0.923} \\

    \bottomrule
\end{tabular}
\caption{\footnotesize{Classification performance report of the different models.\newline The optimised and recorded RF is of 600 trees size with the \textit{uni}-gram\newline  BoW encoding (see details in Fig.\ref{fig:oob} ) }}
\label{table:results}
\end{table}

The results strongly suggest that most models are largely skillful at inferring ICMS standards from the short text provided in the BoQs. Simpler models, additionally, like the generic TCN architecture with basic fine-tuning outperforms established recurrent architectures at this task. This can also be observed with the MLP of one hidden layer based on a vector space model for language representation coming on top. It is apparent that information is notably embedded in local key features in the examined short text of BoQs, with marginal signal provided contextually. The performance of the Random Forest can in fact emphasise this further; with 600 bootstrapped samples of the original text and a random subset of terms set to 12 ($\log_{2}|V|$, where $|V| = 6045$) to induce each classification tree, the results are very comparable, and even showed to outperform those of recurrent architectures, implying that inference from BoQs is effectively achievable. In fact the choice for the RF was due to their simplicity relative to deep neural networks, and to their ability to handle imbalance (and noise) in data through resampling, and randomised selection of terms in text relative to the multiple induced classification trees. Its performance was nearly as accurate as best performing model on the dataset, only outperformed by the the MLP and TCN. 

\begin{figure*}
  \includegraphics[width=\linewidth]{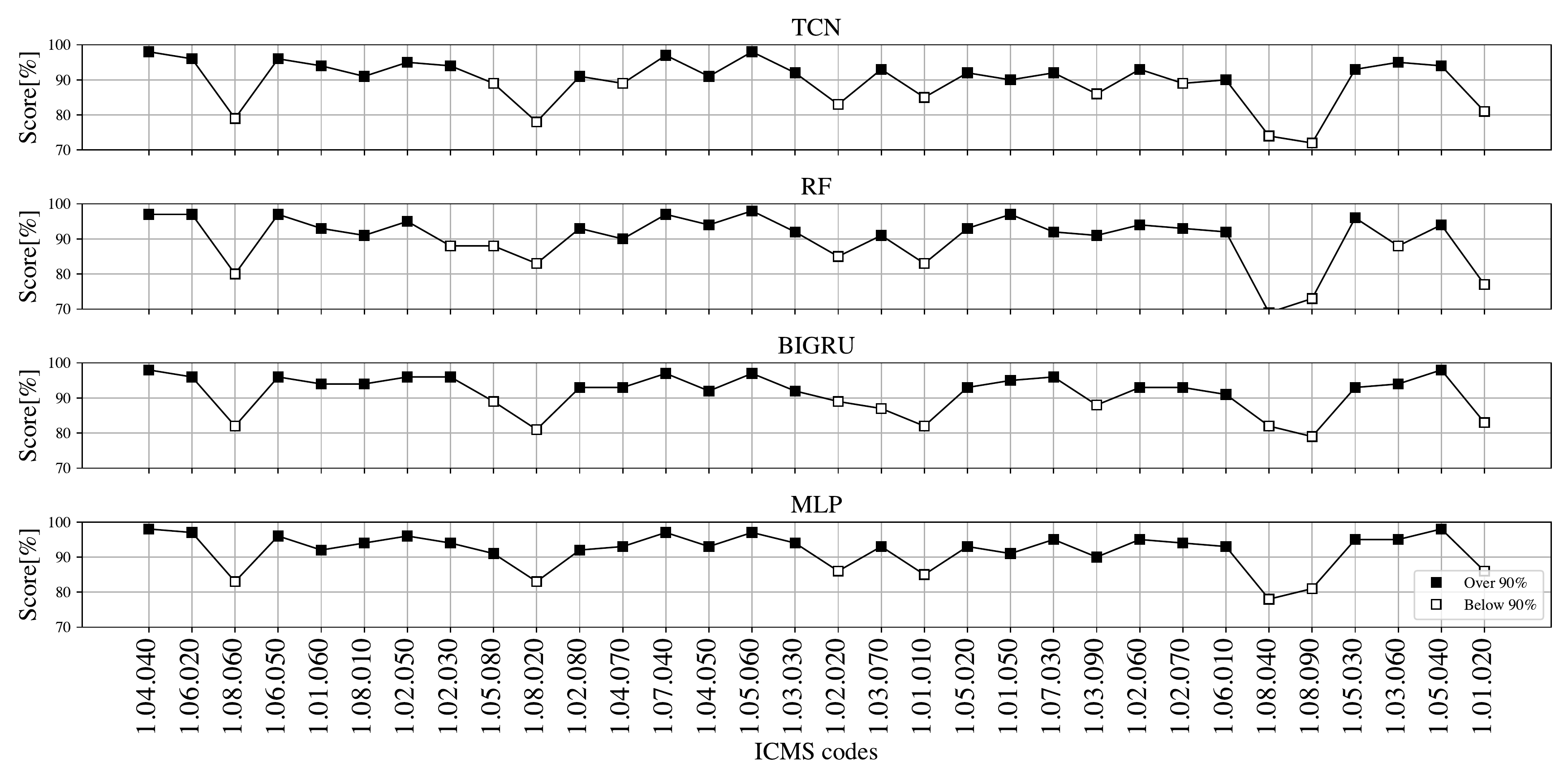}
  \caption{F1 score per ICMS category recorded for the TCN, RF, BiGRU and MLP on the test set. Black dots mean over 90\%, white dots are below 90\% F1 score. MLP (bottom) achieves a $\geq 90\%$ F1 Score on 25 ICMS categories, compared to the rest of models with similar performance on about 22 categories. }
  \label{fig:f1scorebest4}
\end{figure*}

The F1 scores corresponding to each ICMS category recorded on the test set for the best performing models is reported in Fig. \ref{fig:f1scorebest4}. Again, results are closely comparable exhibiting a high performing inference skill overall, with performance on most ICMS classes considered in this study ($\geq 21$ of 32 total) above $90\%$ accuracy. \newline \newline The feed-forward MLP outperforms all models, achieving $\geq 90\%$ F1 score on 25 ICMS categories, which again confirms the suggestion that despite the presence of a semantic component in a subset of BoQs language, simpler models can capture underlying structure from key terms pertinent in their predominantly descriptive text. 

The classification performance on the three ICMS classes below\footnote{For a full list of ICMS definitions reader is advised to refer to \cite{mitchell2016international}.}:
\begin{itemize}
    \item $1.08.060$ : ``Preliminaries | Constructor's site overheads | general requirements: Other temporary facilities and services"
    \item $1.08.020$ : ``Preliminaries | Constructor's site overheads | general requirements: Temporary access roads and storage areas, traffic management and diversion (at the Constructors’ discretion)"
    \item $1.05.080$ : ``Services and equipment: Control systems and instrumentation"
\end{itemize}
was lower than expected given the amount of training samples relative to these categories in the dataset (see Fig. \ref{fig:histoICMS}). Albeit still reasonably accurate, as models generally achieved above $80\%$ F1 score, they under-performed on these categories with respect to their overall (mean) performance. We observed that many of these classes were either of a broader nature (e.g., $1.08.060$) or seemed more context-dependent (e.g., $1.08.020$) or both broad and context-dependent as in $1.05.080$, which additionally included samples with considerable reference to equipment-related jargon and services with varying levels of detail; from descriptions of one word length like ``\textit{spares}'', ``\textit{cabinet}'', ``\textit{CCTV}'' to more informative descriptions like ``\textit{Loop detector installation type [number] lane in main carriageway at [location] [number]}'' and ``\textit{Following a detailed assessment of [location] and sight lines to the new Entry Slip Signals a number of changes will need to be made to the entry slips of the scheme}''. 

Many of these descriptions additionally overlapped across the different codes, for example the terms ``\textit{CCTV}'' and ``\textit{control room/site}'' used in nearly the same wording appeared frequently in both $1.08.020$ and $1.08.060$ categories, which illustrates how complex, if at all possible, is to classify these items, especially where the difference lays in contextual information which is not provided. \newpage Here, a \textit{permanent} CCTV is presumably of one class, and a \textit{temporary} one is of another, whereas neither the words ``\textit{permanent}'' nor ``\textit{temporary}'' were necessarily present.  That is, in order to improve inference beyond this point more training data of diverse contextual nature has to be provided, and subsequently modelled.

On the other hand, the under-performance of the different models observed on 4 to 7 categories consistently below \%90 F1 score is partly caused by the same reasons stated earlier, amplified by the long-tailed distribution of samples across the 32 ICMS standards considered in the dataset. The majority of these categories happen to be significantly under-represented in the original dataset, and many of them stand only slightly above the 250-samples cutoff which was applied. This is to be compared to the mean of about 1600 samples per class in the dataset. Highly skewed datasets, where the minority classes are heavily outnumbered by one or more classes, have proven to be a challenge while at the same time becoming more and more common \cite{DA}. A conservative solution to this conundrum has been to under-sample by deleting the very minority classes as done with classes of less than $250$ samples. Although we applied this limit relatively arbitrarily, it has been set as a trade-off between the classification of a larger number of ICMS categories on the one hand,  and model stability on another. \newline Alternatively in absence of richer (and potentially larger) datasets, methods for data augmentation in NLP (e.g., token-level perturbation like EDA \cite{EDA}, misclassified samples augmentation \cite{Counterexample} and techniques for under and oversampling like SMOTE \cite{chawla2002smote} and MLSMOTE\cite{MLSMOTE}, among others), which have shown improved performance on many text classification tasks, could be potentially applied here\footnote{For a comprehensive review of data augmentation methods in NLP reader is advised to refer to \cite{DA}.}. In this study however, only the bootstrapping of the random forest was applied as overall classification performance was largely up to the mark.

\begin{figure}[ht]
  \includegraphics[width=\linewidth]{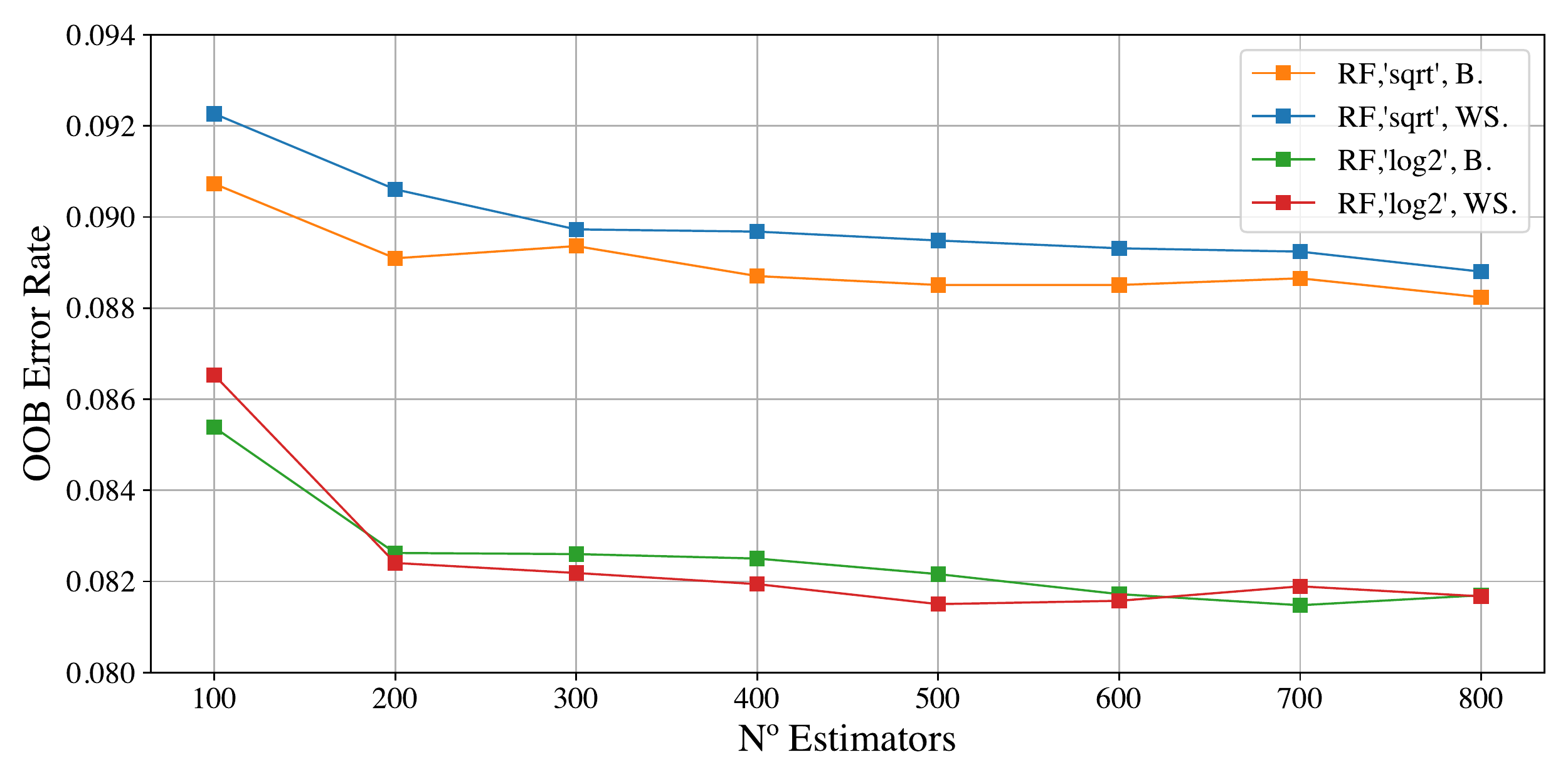}
  \caption{The out-of-bag (OOB) of the different RF configurations on the development set. The models with less number of random features at each classification tree, corresponding to $\log_2 |V|$, show better skill at the inference task. Similarly a slight improvement can be achieved when the \textit{warm-start} (WS) hyper-parameter is activated. WS is a parameter provided by some Scikit-models to allow existing fitted model attributes to initialise a new model in a subsequent call to fit.}
  \label{fig:oob}
\end{figure}

\begin{figure}[ht]
  \includegraphics[width=\linewidth]{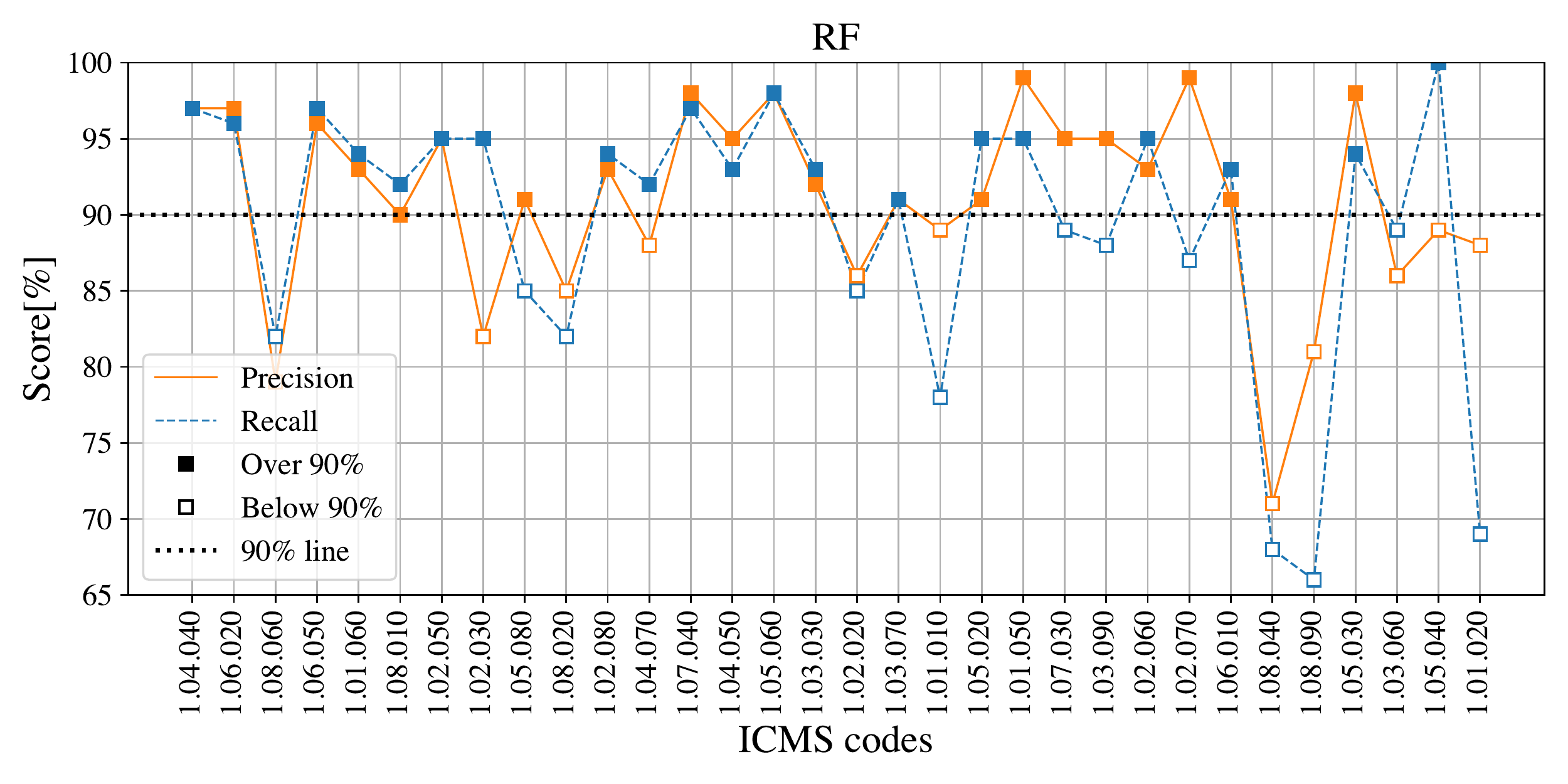}
  \caption{Precision and Recall scores of the RF on the 32 ICMS categories. Although the average precision is very high, the algorithm is more sensitive to small data samples than the neural network -based models. Compare to figure \ref{fig:f1scorebest4}}
  \label{fig:RF-PR}
\end{figure}

All models where tuned and optimised experimentally. The reported performance of the SVM and Multinomial NB correspond to their best models tuned with cross-validation (K-fold) on the development set. For the Random Forest we tuned the models on the development set to minimise the estimated \textit{out-of-bag} (OOB)\cite{oob} error as provided in Fig. \ref{fig:oob}, which showed noticeable convergence of performance towards a size of 600 classification trees. We additionally report both the Precision and Recall scores corresponding to each ICMS category recorded on the test set for the optimal RF in Fig.\ref{fig:RF-PR}, separately. \footnote{The performance of RF of 600 trees is reported here. We provide access to the trained model in production alongside the implementation of the evaluated models in this study.} \newpage As was foreseeable this again shows some arguably peripheral under-performance of the model on instances of under-represented categories as described earlier. Despite a better performance that has been achieved on these particular samples by the different deep learning models, compared to the RF, it can be overenthusiastic to draw conclusive arguments as to why that was, nonetheless.

\begin{figure} [ht!]
  \includegraphics[width=\linewidth]{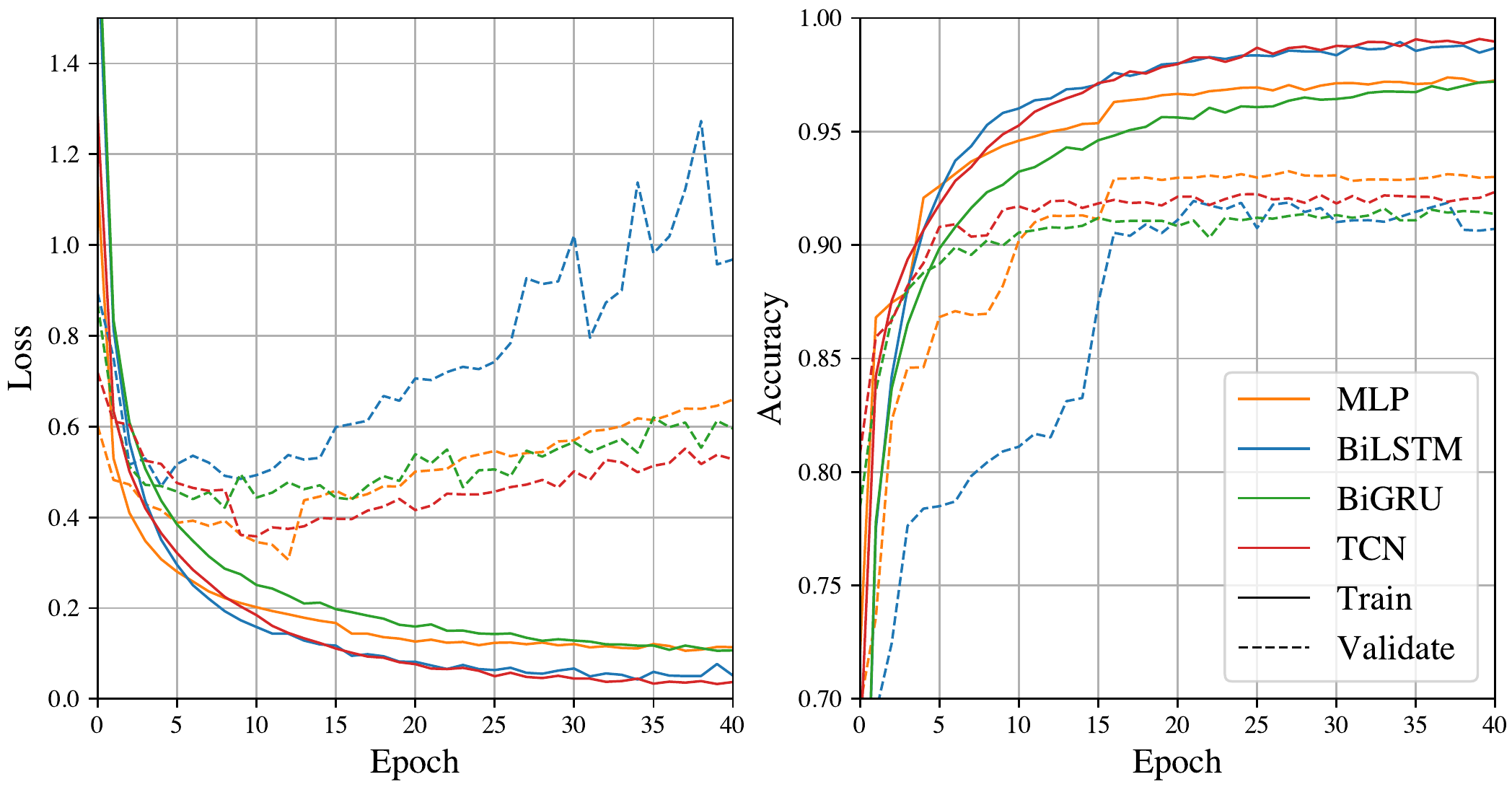}
  \caption{Learning, validation and loss performance for the different ANN architectures on the BoQ dataset. Whilst all models learning converges quickly in training over the first 10 epochs, approximately asymptotic thereafter, the TCN and MLP seem to exhibit less variance with better accuracy performance overall. Recurrent architectures, and especially the BiLSTM, appear to have highest variance on the dataset as a result of learning a much larger set of parameters.}
  \label{fig:MLP}
\end{figure}

Different configurations for the different ANN models used in this study were evaluated. The best model was saved and their performance was reported on the test set. The criteria for initial selection as candidate options included their reported performance in a wide range of language processing applications and benchmarks, whereas the architecture parameters where optimised relative to the classification performance of the models as well as that of their learning. Learning rate and drop-out rates were fixed as reported earlier. Most models showed similar learning performance (and loss minimisation rate) over the training epochs as shown in Fig. \ref{fig:MLP}, and were able to converge at 15 to 20 training epochs. Though again more complex models, e.g., BiLSTM, whilst converging nearly similarly to the rest of models, seem more prone to over-fitting, exhibiting considerable difference between training and validation loss over the successive learning process, and are as such sub-optimally adjusted.

There is a marginal improvement in performance as word embedding is learned by the models over using pre-trained vectors, although at some computational cost, as models using pre-trained vectors were relatively faster to train. Consistently nonetheless these models showed higher loss on validation instances. Although the corpus used to train models can be deemed sufficiently large, quantitatively, it's however less so semantically, especially due to its descriptive and short nature, and the considerable presence of specialised, and occasionally non-English, language. This can explain the marginal edge achieved by learning an embedding vector for language representation on this corpus compared to a pre-trained one, consistent with the conclusions in \cite{embedding_construction} on the benefits of learning word embeddings for the construction domain.  

In general, the experimental results indicate an effective high inference skill of all ANN architectures on this task, with comparable results additionally available with RFs. In fact due to the specific nature of language use in the BoQs; short, descriptive and technical, simpler models achieved better accuracy performance. Both MLP and TCN showed to be able to outperform other---more sophisticated---methods.
As mentioned earlier, the position in a text is only important once the context is inferred from the text. In the case of short texts, context simply isn't provided, and can only be inferred by experts by looking at other variables or based on previous knowledge of the project, most of which is not modelled. The “more-flexible-memory” advantage of RNNs is therefore largely inconsequential at this task, and as a result the TCN exhibited comparable memory to recurrent architectures with the same capacity. It also has a very small number of parameters compared to the BiGRU and BiLSTM networks, and as the texts are too simple to make use of this added complexity, these models tend to overfit and comparatively underperform.

\section{Conclusion and Impact}

This work presents the first attempt to automate the (still manually-handled) mapping of free written work and items' cost text descriptions, from construction cost documents called bills of quantities (BoQs), into the International Cost Measurement Standard (ICMS), which will enable benchmarkers to compare and benchmark the performance of projects at a scale that was never done before, and facilitate more effective cost and risk analysis in construction projects.  To that end we evaluated state-of-the-art machine learning methods to learn multi-class text classification models from 51906 item descriptions. These were retrieved from 24 different infrastructure construction projects carried out by contractors of public-owned companies of the United Kingdom, across the UK. 

We considered two approaches to our modelling, one assuming information signals can be captured from local features of the description text provided in the BoQs, and another on the premise that, alongside local key features, the potential propagation of information and semantics in the text may help improve the learning of ICMS codes. To do that we evaluated a range of classification methods which have been widely used on tasks of text classification, including support vector machines, random forests, multi-layer perceptron, and advanced deep learning architectures commonly used in sequence modelling, including recurrent (LSTM, GRU) and convolutional architectures (CNN, TCN). 

Whilst results strongly suggest that most models are largely skillful at inferring ICMS standards from the short text provided in the BoQs, we found that simpler models, like the RF, and generic MLP and TCN architectures with minimal tuning outperform recurrent --more sophisticated-- architectures such as LSTMs and GRUs. This is likely due to the ``straight-to-the-point'' nature of text found in the BoQs. That is, they are considerably condensed, short and strictly descriptive, so much so that their complexity strikes as being a function of abstraction in key term use, rather than the inherent complexity of semantic dynamics in language use more often than not. The “long memory” advantage of RNNs is therefore largely inconsequential at this task, and as a result the TCN exhibited comparable memory to recurrent architectures with the same capacity. And simpler models like the MLP and RF were able to capture the required mapping favourably from local key features. 

As adoption of ICMS gains traction, more annotated data will be made available and the evaluated models can be re-trained to learn further ICMS categories. It is therefore hoped that the findings of this study will trigger this process further. To that end, the trained model of MLP and development code in this study are made available to the community and can be readily used \footnote{Operational model (MLP) and development code are available on: https://github.com/ignaciodeza/BoQ-classifier-ICMS}. Consequently, we believe this study presents a compelling case for the community -- both private and public sectors -- of the construction industry to prioritise an open data approach along their supply lines, apace with considerable use of tools to ensure friction-less standardisation. We argue this will allow for vital developments in the field leading to a transformative automated benchmarking system. 

\section*{Acknowledgements}
This work has been supported by Innovate UK under Grant N: 08027517 as a part of ``Transport infrastructure efficiency strategy living labs'' (TIES Living Labs) Project N. 106171.

\bibliographystyle{elsarticle-num}
\bibliography{ICMS_ML}

\end{document}